\def\year{2020}\relax
\documentclass[letterpaper]{article} 
\usepackage{aaai20}  
\usepackage{times}  
\usepackage{helvet} 
\usepackage{courier}  
\usepackage[hyphens]{url}  
\usepackage{graphicx} 
\usepackage{graphicx}  
\usepackage{amsmath}
\usepackage{booktabs}
\usepackage{algorithm}
\usepackage{algorithmic}
\usepackage{subcaption}
\usepackage{amssymb}
\usepackage{bm}
\usepackage{bbm}
\usepackage{multirow}
\usepackage{siunitx}

\title{Efficient Automatic CASH via Rising Bandits}
\author{Yang Li,\textsuperscript{\rm 1} Jiawei Jiang,\textsuperscript{\rm 2} Jinyang Gao,\textsuperscript{\rm 3} Yingxia Shao,\textsuperscript{\rm 4} Ce Zhang,\textsuperscript{\rm 2} Bin Cui\textsuperscript{\rm 1}\\ 
\textsuperscript{\rm 1}Key Laboratory of High Confidence Software Technologies (MOE), School of EECS, Peking University, Beijing, China\\
\textsuperscript{\rm 2}Department of Computer Science, Systems Group, ETH Zurich, Switzerland\\ 
\textsuperscript{\rm 3}Beijing University of Posts and Telecommunications, Beijing, China\\
\textsuperscript{\rm 4}Alibaba Group, Hangzhou, China\\
\{liyang.cs, bin.cui\}@pku.edu.cn,  \{jiawei.jiang, ce.zhang\}@inf.ethz.ch\\
jinyang.gjy@alibaba-inc.com, shaoyx@bupt.edu.cn
}
 
\begin{document}

\maketitle

\begin{abstract}
The Combined Algorithm Selection and Hyperparameter optimization (CASH) is one of the most fundamental problems in Automatic Machine Learning (AutoML). 
The existing Bayesian optimization (BO) based solutions turn the CASH problem into a Hyperparameter Optimization (HPO) problem by combining the hyperparameters of all machine learning (ML) algorithms, and use BO methods to solve it. 
As a result, these methods suffer from the low-efficiency problem due to the huge hyperparameter space in CASH.
To alleviate this issue, we propose the alternating optimization framework, where the HPO problem for each ML algorithm and the algorithm selection problem are optimized alternately. 
In this framework, the BO methods are used to solve the HPO problem for each ML algorithm separately, incorporating a much smaller hyperparameter space for BO methods.
Furthermore, we introduce \emph{Rising Bandits}, a CASH-oriented Multi-Armed Bandits (MAB) variant, to model the algorithm selection in CASH.
This framework can take the advantages of both BO in solving the HPO problem with a relatively small hyperparameter space and the MABs in accelerating the algorithm selection.
Moreover, we further develop an efficient online algorithm to solve the Rising Bandits with provably theoretical guarantees.
The extensive experiments on $30$ OpenML datasets demonstrate the superiority of the proposed approach over the competitive baselines.

\end{abstract}

\section{Introduction}

Machine learning (ML) has made great strides in many application areas, e.g., recommendation, computer vision, financial market analysis, etc~\cite{goodfellow2016deep,he2017neural,ma2019mmm,zhao2019dmdp}.
However, given a practical application, it is usually knowledge-intensive and labor-intensive to develop customized solutions with satisfied learning performance, where the exploration may include but is not limited to selecting ML algorithms, configuring hyperparameters and network architecture searching.
To facilitate the deployment of ML applications and democratize the usage of machine learning, it is of vital importance to reduce human efforts during such exploration.
Naturally, automatic machine learning~\cite{quanming2018taking,DBLP:journals/corr/abs-1904-12054} has attracted lots of interest from both industry and academia.

Given a learning problem, the first thing is to decide which ML algorithm should be applied -- from SVM, Adaboost, GBDT \cite{jiang2018dimboost,jiang2017tencentboost} to deep neural networks.
According to the No Free Lunch theorem~\cite{ho2001simple}, no single ML algorithm can achieve the best performance for all the learning problems; and there is often no golden standard to predict which ML algorithm performs the best.
As a result, we typically spend computational resources across all reasonable ML algorithms, and choose the one with the best performance after the optimization of their hyperparameters and network architectures.
However, solving the algorithm selection problem after sufficiently optimizing the hyperparameters of each ML algorithm leads to inefficient usage of computational resources.
Resources consumed by the poor-performing algorithms are greatly wasted.
To this end, the Combined Algorithm Selection and Hyperparameter optimization (CASH) problem~\cite{feurer2015efficient,kotthoff2017auto} is proposed to jointly optimize the selection of algorithm and its hyperparameters, which is the core focus of this paper.

To solve the CASH problem, a class of methods~\cite{komer2014hyperopt,feurer2015efficient,kotthoff2017auto} transform the CASH problem into a unified hyperparameter optimization (HPO) problem by merging the hyperparameter space for all ML algorithms and 
treating the selection of algorithm as a new hyperparameter.
Then classical Bayesian optimization (BO) methods~\cite{shahriari2015taking} are utilized to solve this HPO problem.
Consequently, these methods incorporate a huge optimization space with high-dimensional hyperparameters for BO methods.
Past works~\cite{eggensperger2013towards} show that BO methods perform well for relatively low-dimensional hyperparameters.
However, for high-dimensional problems, standard BO methods perform even worse than random search~\cite{wang2013bayesian}.
Thus, such a huge hyperparameter space greatly hampers the efficiency of Bayesian optimization.

To alleviate the above issue, it is natural to consider another paradigm where the BO methods are used to solve the HPO problem for each ML algorithm separately, and the algorithm selection is responsible for determining the allocation of resources to each ML algorithm's HPO process. 
Based on this idea, we propose the \emph{alternating optimization framework}, where the HPO problem for each ML algorithm and the algorithm selection problem are optimized alternately.
Benefiting from solving the HPO problem for each ML algorithm individually, this framework brings a much smaller hyperparameter space for BO methods.
Furthermore, within this framework, the resources can be adaptively allocated to the HPO process of each algorithm based on their performance.
Intuitively, spending too many resources in tuning the hyperparameters of poor-performing algorithms should be avoided; instead, more resources should be allocated to the more promising ML algorithms that can achieve the best performance. 
Unfortunately, which algorithm is the best is unknown unless enough resources are allocated to its HPO process. 
Therefore, solving the CASH problem efficiently requires to trade off the well-celebrated Exploration vs. Exploitation (EvE) dilemma during algorithm selection: \emph{should we explore the HPO of different ML algorithms to find the optimal algorithm (Exploration), or give more credit to the best algorithm observed so far to further conduct HPO (Exploitation)?}

Since the EvE dilemma has been intensively studied in the context of Multi-Armed Bandits (MAB), here we propose to solve the algorithm selection problem in the framework of MAB.
In this setting, each arm is associated with the corresponding HPO process of an ML algorithm.
Pulling an arm means that a unit of resource is assigned to the HPO process of the corresponding algorithm, and the reward corresponds to the result from the HPO process.
However, the existing MABs cannot be directly applied to model the algorithm selection problem for two reasons:
1) the well-studied objectives of MABs (e.g., accumulated rewards) are not consistent with the target of CASH problem that aims to maximize the observed reward;
2) because the HPO results will be improved with the increase of the HPO resource, the reward distribution of each arm is not stationary over time.

The main contributions of this paper are the following:
\begin{itemize}
    \item We propose the alternating optimization framework to solve the CASH problem efficiently, which optimizes the algorithm selection problem and the HPO problem for each ML algorithm in an alternating manner. 
    It takes the advantages of both BO methods in solving the HPO problem with a relatively small hyperparameter space and MABs in accelerating the algorithm selection.
    \item We introduce a novel, CASH-oriented MAB formulation, termed \emph{Rising Bandits}, where each arm's expected reward increases as a function of the number of times it has been pulled. To the best of our knowledge, this is the first work that models the algorithm selection problem in the framework of non-stationary MABs.
    \item We present an easy-to-follow online algorithm for the Rising Bandits, accompanied with provably theoretical guarantees.
    \item The empirical studies on $30$ OpenML datasets demonstrate the superiority of the proposed method over the state-of-the-art baselines in terms of final accuracy and efficiency. 
    Our method can achieve an order of magnitude speedups compared with BO based solutions.
\end{itemize}

\section{Preliminaries and Related Works}
We first introduce the basic notations for the CASH problem.
There are $K$ candidate algorithms $\mathcal{A}=\{A^1, ..., A^K\}$. 
Each algorithm $A^i$ has a corresponding hyperparameter space $\Lambda_i$. 
The algorithm $A^i$ with a hyperparameter $\lambda$ is denoted by $A^i_{\lambda}$. 
Given the dataset $D=\{D_{train}, D_{valid}\}$ of a learning problem, the CASH problem is to find the joint algorithm and hyperparameter configuration $A^{\star}_{\lambda^{\star}}$ that minimizes the loss metric (e.g., the validation error on $D_{valid}$):
\begin{equation}
    A^\ast_{\lambda_\ast} = \operatornamewithlimits{argmin}_{A^i\in\mathcal{A},\lambda\in\Lambda^i} \mathcal{L}(A_\lambda^i, D).
\end{equation}

Hyperparameter optimization (HPO) is to find the hyperparameter configuration $\lambda^{\star}$ of a given algorithm $A^i$, which has the best performance on the validation set,
\begin{equation}
    \lambda^{\star} = \operatorname{argmin}_{\lambda \in \Lambda_i} L(A^i_{\lambda}, D).
\end{equation}
Bayesian optimization (BO) has been successfully applied to solve the HPO problem.
Sequential Model-based Algorithm Configuration (SMAC)~\cite{hutter2011sequential}, Tree-structure Parzen Estimator (TPE)~\cite{bergstra2011algorithms}, and Spearmint~\cite{snoek2012practical} are three well-established methods. 
It is important to note that these approaches can be executed in a sequential manner. That is, the HPO process is iterative.
Recently, many approaches develop some elaborated mechanisms to allocate the HPO resources adaptively~\cite{huang2019efficient,falkner2018bohb,sabharwal2016selecting}.
In addition, multi-fidelity optimization has been deeply studied in the framework of  BO to accelerate the HPO problem~\cite{swersky2013multi,klein2017fast,kandasamy2017multi,poloczek2017multi,hu2019multi}.

In the algorithm selection problem, the objective is to choose a parameterized algorithm $A^{\star}_{\lambda^{\star}}$, which is the most effective with respect to a specified quality metric $Q(.)$. 
This sub-problem can be stated as a minimization problem:
\begin{equation}
    A_{\lambda^{\star}}^{\star} = \operatorname{argmin}_{i \in [1,...,K]} Q(A_{\lambda^{\star}}^i, D).
\end{equation}
In practice, all candidate algorithms with some fixed hyperparameters are evaluated on the validation dataset, and the algorithm with the best performance is chosen.
However, this method suffers from the ``low accuracy" issue due to the lack of the HPO: the fixed hyperparameters cannot accurately reflect the performance of the algorithm across different problems.
Moreover, many methods select algorithms according to some theoretical decision rules, meta-learning methods~\cite{abdulrhaman2015algorithm} and supervised learning techniques~\cite{sun2013pairwise}.

To solve the CASH problem effectively in the ML applications, it is necessary to select the algorithm and its hyperparameters simultaneously.
Auto-Weka is the first work devoted to the CASH problem, which takes the BO based solutions. 
Then Auto-Sklearn and Hyperopt-Sklearn also adopt the same BO based framework.
In addition, tree-based pipeline optimization tool (TPOT)~\cite{olson2019tpot} uses genetic programming to address the CASH problem.
Recently, Reinforcement learning method~\cite{efimova2017fast} and MAB based methods~\cite{DBLP:journals/corr/abs-1905-00424} have been studied to solve the CASH problem.
They model the rewards in the stationary environment and ignore the objective's difference between MABs and CASH. 
In the community of MAB, several works~\cite{besbes2014stochastic,jamieson2016non,heidari2016tight,levine2017rotting} focus on the non-stationary bandits, but none of them match the settings in CASH.

\section{The Proposed Method}
In this section, we introduce the alternating optimization framework, give the formulation of \emph{Rising Bandits}, and describe the online algorithm to solve this bandit problem.

\subsection{The Alternating Optimization Framework}
We reformulate the CASH problem into the following bilevel optimization problem:
\begin{equation}
    \begin{aligned}
    & \operatornamewithlimits{min}_{i\in[1,...,K]}
    & & Q(A^i_{\lambda^\ast}, D)\\
    & \text{ s.t.}
    & & \lambda^\ast = \operatorname{argmin}_{\lambda\in\Lambda_{i}} L(A^i_{\lambda}, D).
    \end{aligned}
\end{equation}
Here the CASH problem is decomposed into two kinds of sub-problems: algorithm selection problem (the upper-level sub-problem) and the HPO problem for each ML algorithm (the lower-level sub-problem). 
We propose to solve this bilevel optimization problem by optimizing the upper-level and lower-level sub-problems alternately. 
We name it the \emph{alternating optimization framework}.
In this framework, Bayesian Optimization (BO) methods are used to conduct HPO for each ML algorithm individually; MAB based method is utilized to solve the algorithm selection problem. 
This framework brings two benefits:
\begin{itemize}
    \item Since the hyperparameter space for each ML algorithm is relatively small, BO methods can solve the corresponding HPO problem efficiently.
    \item The resources can be adaptively allocated to the HPO of each ML algorithm according to its HPO performance in the MAB framework.
\end{itemize}
As a result, the poor-performing ML algorithms will be equipped with few HPO resources (e.g., the number of trials), and more resources are allocated to the promising algorithms that can achieve better learning performance.

\subsection{Non-stationary Rewards from Bayesian Optimization}
\begin{figure}[t]
	\centering
		\includegraphics[width=0.8\columnwidth]{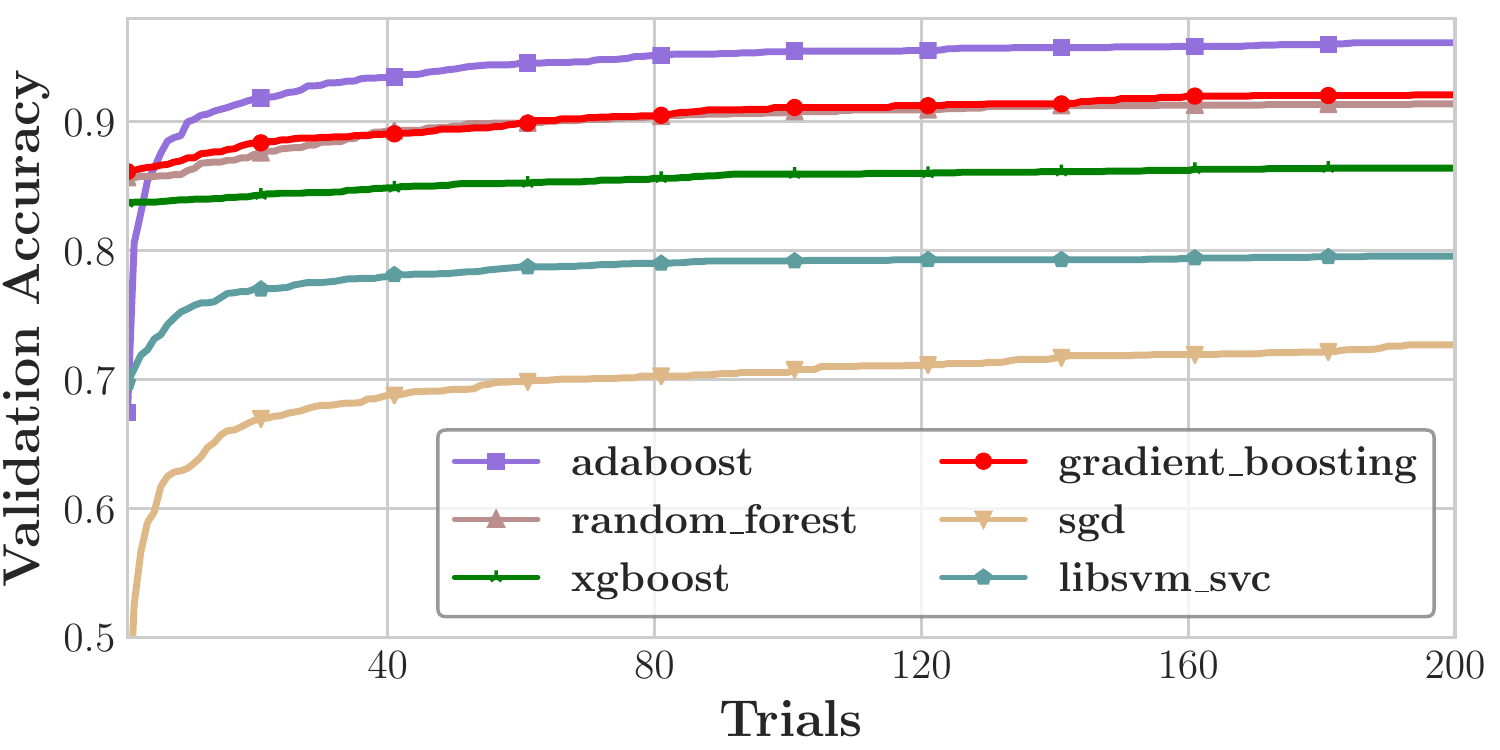}
	\caption{The HPO results of 6 ML algorithms. BO method -- SMAC is used to tune the hyperparameters of each algorithm 50 times, and the average validation accuracy across trials is reported.}
    \label{non_stationary_rewards}
\end{figure}

Before introducing the Rising Bandits, we first investigate the rewards (HPO results) from BO methods.
Given more HPO resources, the expected rewards (i.e., the best-observed validation accuracy) will increase.
Figure \ref{non_stationary_rewards} provides an intuitive example. 
Six ML algorithms are equipped with 200 trials to conduct HPO. 
The rewards $r(.)$ correspond to the best-observed validation accuracy in each trial.
As the number of HPO trial increases, this validation accuracy improves gradually, and then gets saturated.
Further, we can summarize the following observations about the rewards from BO:
\begin{itemize}
    \item For each ML algorithm $A^k$, the reward sequence $r_k(1), ..., r_k(n)$ is increasing and bounded, and the limit $\operatorname{lim}_{n\rightarrow{\infty}}r_k(n)$ exists.
    \item The reward sequence satisfies the \emph{decreasing marginal returns} approximately. Here we abuse the terminology and refer to this as ``concavity''.
\end{itemize}
Since the rewards increase monotonically across trials, it is evident that the rewards are not identically distributed, but are generated by a non-stationary stochastic process.

\subsection{The Definition of Rising Bandits}
Based on the observations about the HPO results, we give the formulation of Rising Bandits to model the algorithm selection problem with non-stationary rewards. 
In this bandit variant, the agent is given $K$ arms, and at each time step $t=1,2,...,T$ one of the arm must be pulled. 
Each arm $k$ is associated with the HPO process of an ML algorithm $A^{k}$. 
Pulling an arm means that a unit of resource (e.g., an HPO trial) is assigned to the HPO process of an algorithm, and the reward corresponds to the non-stationary HPO results. 

In Rising Bandits, we model the non-stationary reward sequences of the arms as follows: each
arm $k$ has a fixed underlying reward function denoted by $r_k(.)$. 
All the reward functions are bounded within $[0, 1]$.
When the agent pulls arm $k$ for the $n^{th}$ time,
he receives an instantaneous reward $r_k(n)$.
We denote the arm that is pulled at time step $t$ as $i(t) \in [K] = [1,...,K]$. 
Let $N_k(t)$ be the number of pulls of arm $k$ at time step $t$, not including this round's choice, that's, $N_k(1)=0$, and $\Pi$ the set of all sequence $i(1), i(2), ...$, where $i(t) \in [K], \forall{t} \in \mathbbm{N}$. 
i.e., $\pi \in \Pi$ is a sequence of actions (arms), also referred to as a policy.
We denote the arm that is chosen by policy $\pi$ at time step $t$ as $\pi(t)$. 

Instead of maximizing the accumulated rewards $\sum_{t=1}^{T}r_{\pi(t)}(N_{\pi(t)}(t)+1)$, the objective of the agent in CASH is to maximize the observed reward within $T$, defined for policy $\pi \in \Pi$ by,
\begin{equation}
    \label{ori_obj}
    J(T;\pi) = \max\limits_{t=1:T} r_{\pi(t)}(N_{\pi(t)}(t) + 1).
\end{equation}
We consider the equivalent objective of minimizing the regret within $T$ defined by,
\begin{equation}
    R(T;\pi) = \max\limits_{\tilde{\pi} \in \Pi}\{J(T;\tilde{\pi})\} - J(T; \pi).
\end{equation}

Based on the observations about the non-stationary rewards, we introduce the following assumption: \newline
\textbf{Assumption 1. }
\emph{\textbf{(Rising)} $\forall{k} \in [K]$, $r_k(n)$ is bounded, increasing, and concave in $n$}. 
\newline
According to this assumption, the original objective in (\ref{ori_obj}) is equivalent to, 
\begin{equation}
     J(T;\pi) = \max\limits_k r_k(N_k(T+1)).
\label{cash_obj}
\end{equation}
In the CASH problem, it is clear that the reward function $r(n)$ is bounded and increasing; but the concavity assumption may not always hold. 
We will discuss the two situations in the following sections.
Then we investigate an offline solution for the Rising Bandits. 
The offline setting means that the optimal arm is known to the agent before the game.
Let $\pi^{max}$ be a policy defined by,
\begin{equation}
    \pi^{max}(t) \in \operatornamewithlimits{argmax}_{k \in [K]}{r_k(T)}.
\end{equation}
\textbf{Lemma 1.} 
\emph{
$\pi^{max}$ is the optimal policy for the Rising Bandits problem in the offline setting. \newline
\textbf{Proof}: See Appendix A.1 of the supplementary material.
}
\newline
If the best arm is known to the agent, the optimal policy must pull the best arm repeatedly.

\subsection{Online Solution for Rising Bandits}
The CASH problem falls into the online setting, where the best arm is unknown to the agent. 
In this circumstance, the above Lemma $1$ fails.
However, it guides us to derive an efficient policy in the online setting: 1) first identify the best arm by using as few time steps as possible, and then 2) pull the best arm until the time step $T$ meets.
That is, solving the best arm identification problem first and then fully exploiting the best arm can efficiently optimize the objective in (\ref{cash_obj}). 

Based on the Assumption 1, we can obtain an interval that bounds the final reward of an arm.
The reward function is concave, that's, for any $n>2$, we have $r(n) - r(n-1) \ge r(n+1) - r(n)$.
Suppose the arm $k$ has been pulled $n$ times, and $n$ rewards $r_k(1),..., r_k(n)$ are observed.
Given that $r_k(.)$ is increasing, bounded and concave, we have for any $t > n$,
\begin{equation}
\label{ub}
    r_k(t) \le \operatorname{min}(r_k(n) + (t - n) \omega_k(n), 1),
\end{equation}
where $\omega_k(n)$ equals $r_k(n) - r_k(n - 1)$, and we name $\omega(n)$ as the growth rate at the $n^{th}$ step. 
We refer to the right-hand side of Inequality \ref{ub} as the upper bound $u_k(t)$ of $r_k(t)$. 
Naturally, the lower bound $l_k(t)$ of $r_k(t)$ is $r_k(n)$. 
If the arm $i$ has the lower bound $l_i(t)$ that is no less than the upper bound $u_k(s)$ of the arm $k$, the arm $k$ cannot be the optimal arm. 
By using this elimination criterion, we can gradually dismiss the arm that cannot be the optimal arm. 
After finding the best arm, this arm will be pulled repeatedly until the game ends.

Algorithm \ref{algo:algo_bandit} illustrates both the pseudo-code of the proposed online algorithm and its usage in the alternating optimization framework.
It operates as follows: it maintains a set of candidate arms (ML algorithms) in which the best arm is guaranteed to lie (Line 1). 
At each round, it pulls all the arms in the candidate set once, and it means that each corresponding algorithm in the candidate set is given one unit of resource to tune its hyperparameters with BO methods. 
Then both the upper bound and lower bound of the final reward at time step $T$ are updated (Line 5-10). 
If the upper bound of the final reward of an arm $k$ (algorithm $A_{k}$) is less than or equal to the lower bound of another arm's in the candidate set, then the arm $k$ will be eliminated from the candidate set (Line 11-15).
The above process iterates until the time step $T$ meets.
Finally, the best algorithm along with the corresponding hyperparameter configuration is returned.

\begin{algorithm}[tb]
  \caption{Online algorithm for Rising Bandit}
  \label{algo:algo_bandit}
  \textbf{Input}: ML algorithm candidates $\mathcal{A}=\{A_1,...,A_K\}$, the total time steps $T$, and one unit of HPO resource $\hat{b}$.
  
  \begin{algorithmic}[1]
  \STATE Initialize $S_{cand} = \{1, 2, ..., K\}$, $t = 0$.
  \WHILE{$t < T$}
      \FOR{each $k \in S_{cand}$}
          \STATE $t = t+1$.
          \STATE Pull arm $k$ once:  $H_k\leftarrow{\operatorname{Iterate\_HPO}(A_{k}, \hat{b})}$.
          \STATE Calculate $\omega_k(t)$ according to $H_k$.
          \STATE Update $u_k^t(T) = \operatorname{min}(y_k(t) + \omega_k(t)(T - t), 1)$.
          \STATE Update $l_k^t(T) = y_k(t)$.
      \ENDFOR
      \FOR{$i \ne j \in S_{cand}$}
          \IF{$l_i^t(T) \ge u_j^t(T)$}
              \STATE $S_{cand} = S_{cand}\backslash\{j\}$
          \ENDIF
      \ENDFOR
  \ENDWHILE
  \STATE \textbf{return} the corresponding ML algorithm $A^{\star}$ and its best hyperparameter configuration.
\end{algorithmic}
\end{algorithm}

\subsection{Rising Bandits with ``Loose'' Concavity}
As stated previously, the concavity in Assumption 1 may not always hold in the CASH problem.
In this case, the problematic growth rate $\omega_k(t) = r_k(t) - r_k(t-1)$ may lead to a wrong upper bound. 
To alleviate this issue, we propose to use the following growth rate calculated by,
\begin{equation}
    \omega_{k}(t)=\frac{y_k(t) - y_k(t-C)}{C},
\end{equation}
where $C$ is a constant.
Intuitively, by averaging the latest $C$ growth rates, this smooth growth rate is more robust to the case with ``loose'' concavity.
In the next section, we provide theoretical guarantees for the proposed methods.

\section{Theoretical Analysis and Dissussions}

For the coming theorem, we define a quantity that
captures the time steps required to distinguish the optimal arm from the others. 
More precisely, we define $\gamma(T) = \operatorname{max}_k\gamma_k(T)$, where
\begin{equation}
    \gamma_k(T) = \operatorname{arg}\min\limits_{t}\{u_k^t(T) \le l_{k_{\ast}^T}^t(T)\}
    \end{equation}
and $k_{\ast}^T$ is the optimal arm in the given $T$.
Thus $\gamma_k(T)$ specifies the smallest number of time steps needed to pull both arm $k$ and the optimal arm so that the sub-optimal arms can be distinguished from the optimal arm.

We prove that the upper bound of the policy regret of the proposed algorithm exists.\newline
\textbf{Theorem 1.} 
\emph{
Suppose Assumption 1 holds. The proposed algorithm achieves regret bounded by,
\begin{equation}
    R(T; \bar{\pi}) \le r_{k_{\ast}^T}(T) - r_{k_{\ast}^T}(T - (n-1)\gamma(T)).
\end{equation}
\textbf{Proof}: See Appendix A.2 of the supplementary material.
}
\newline
This bound contains a problem-dependent term $\gamma(T)$. If identifying the optimal arm is easier, $\gamma(T)$ will be smaller.

\subsection{Compare with Average Policy}
An intuitive policy $\pi_{uni}$ is to pull each arm $\frac{T}{K}$ times. 
The regret of this policy is,
\begin{equation}
    R(T; \pi_{uni}) = r_{k_{\ast}^T}(T) - \max\limits_k r_k(\frac{T}{K}).
\end{equation}
We now establish the regret connection between the proposed algorithm and the average policy.
\newline
\textbf{Corollary 1.} 
\emph{
When the problem-dependent term $\gamma(T)$ satisfies: $\gamma(T) \le \frac{KT - T}{K(K-1)}$,
the regret of the proposed algorithm will not be worse than the average strategy's.
\begin{equation}
    R(T; \bar{\pi}) \le R(T; \pi_{uni}).
\end{equation}
\textbf{Proof}: See Appendix A.3 of the supplementary material.
}

\subsection{Theoretical Guarantee for ``Loose'' Concavity}
Here we provide a theoretical guarantee for the smooth growth rate. 
For any reward sequence $y_k(1), ...$, we can find a reward function $r_k(.)$ that satisfies the Assumption 1.
At each time step $t$, $r_k(t) \ge y_k(t)$, and they have the same limit. 
We denote the bias between $r_k(.)$ and $y_k(.)$ by $\Delta_k(t) = r_k(t) - y_k(t)$.
\newline
\textbf{Theorem 2.} 
\emph{
If the following condition holds, the proposed algorithm with smooth growth rate can be used to identify the optimal arm without any loss,
\begin{equation}
    \frac{\Delta_k(t)}{\Delta_k(t-C)} \le \frac{T-t}{T-t+C}.
\end{equation}
\textbf{Proof}: See Appendix A.4 of the supplementary material.
}

\subsection{Towards Cost-Aware CASH}
In the previous sections, the limited resource is the number of HPO trials, and here we consider the time $B$ as the limited resource.
Both the algorithm's performance and its HPO evaluation cost in runtime should be taken into consideration.
In CASH, conducting an HPO trial for different ML algorithms usually takes a different time cost. 
For example, for large datasets, training linear models is much faster than the tree-based model such as gradient boosting.
To solve the cost-aware CASH, we develop a variant of Algorithm $1$.
For each ML algorithm, we first compute its average time overhead $c_k$ in each HPO trial;
then we predict the upper bound of the final reward within the given time $B$ by,
\begin{equation}
    u_k^t(B) = r_k(t) + \omega_k \frac{B^{'}}{c_k},
\end{equation}
where $B^{'}$ is the time left, and $\omega_k$ is the growth rate at time $t$.

\subsection{Relationship and Comparison with Previous Works}
Our approach takes an adaptive resource allocation
scheme. From a theoretical perspective, our method is similar, in spirit, to some previous works~\cite{huang2019efficient,falkner2018bohb,sabharwal2016selecting}. 
In addition, one work~\cite{heidari2016tight} also supports concave reward functions. 
Compared with these works, our main contribution is to apply a similar methodology to a new application, i.e., CASH. In the CASH problem, we find some additional structures that we can use, e.g., CASH has the concave structure in the reward function. 
Furthermore, instead of optimizing the accumulated regrets in~\citeauthor{heidari2016tight}, CASH focuses more on identifying the best arm. 
These additional structures allow us to perform significantly better over simply applying these previous approaches to CASH.

Compared with BO based solutions,
our method explicitly reduces the hyperparameter space in the CASH problem by dismissing the poor algorithms successively. 
Without any modification, this method can also be used to solve the regression tasks by mapping the loss into $[0, 1]$.
In addition, the proposed approach can handle the cost-aware CASH; however, the existing solutions for the CASH problem do not take the evaluation cost into consideration.

\section{Experiments and Results}
In the evaluation of the proposed method, we demonstrate its superiority from the following three perspectives:
1) the efficiency compared with the state-of-the-art BO based solutions, 2) the empirical performance compared with all considered baselines in terms of final accuracy and efficiency, and 3) practicability and effectiveness in the AutoML systems.

\begin{figure*}[t]
    \centering
    \begin{subfigure}[b]{\textwidth}
        \centering
        \includegraphics[width=0.32\columnwidth]{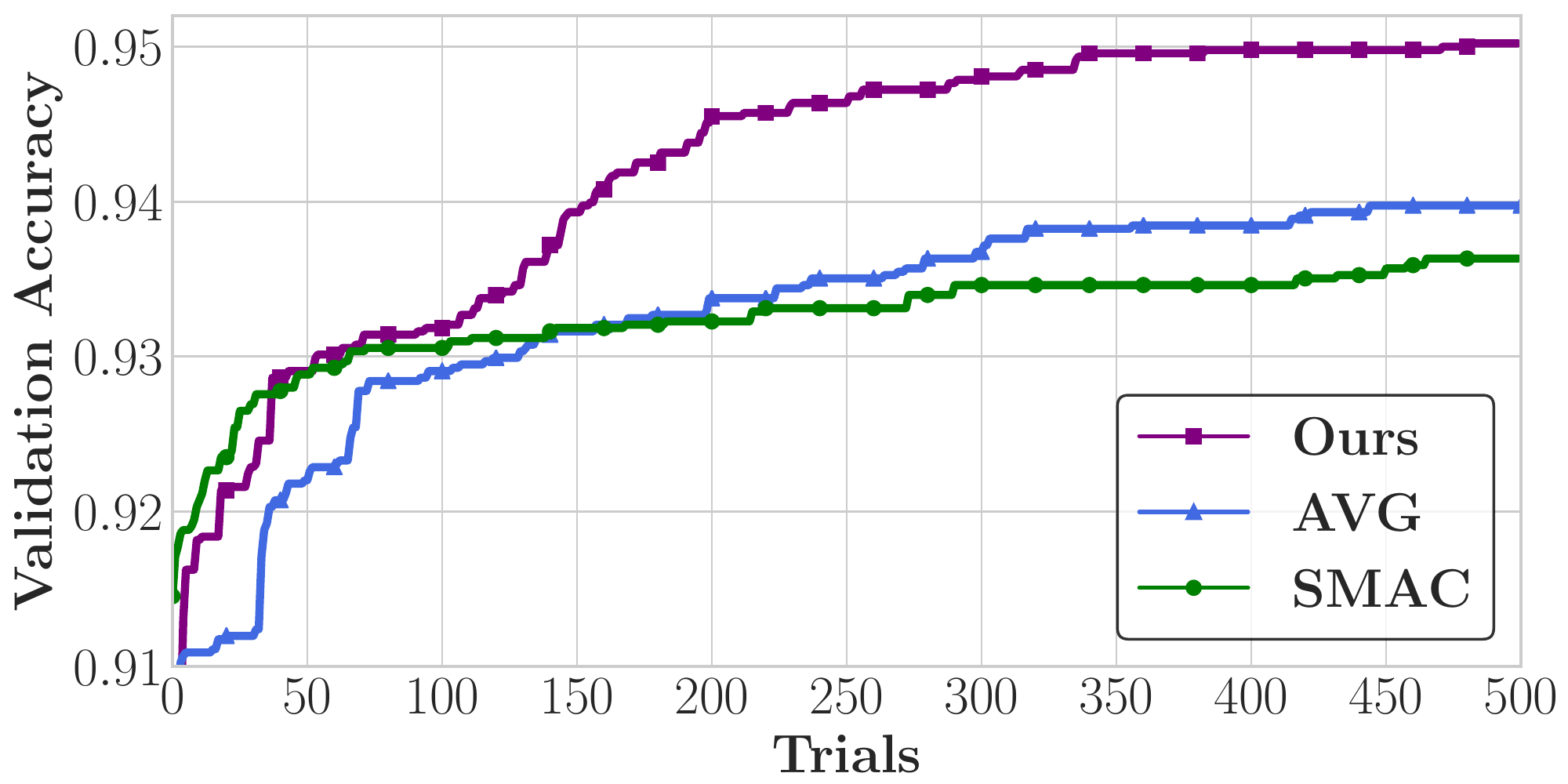}
        \includegraphics[width=0.32\columnwidth]{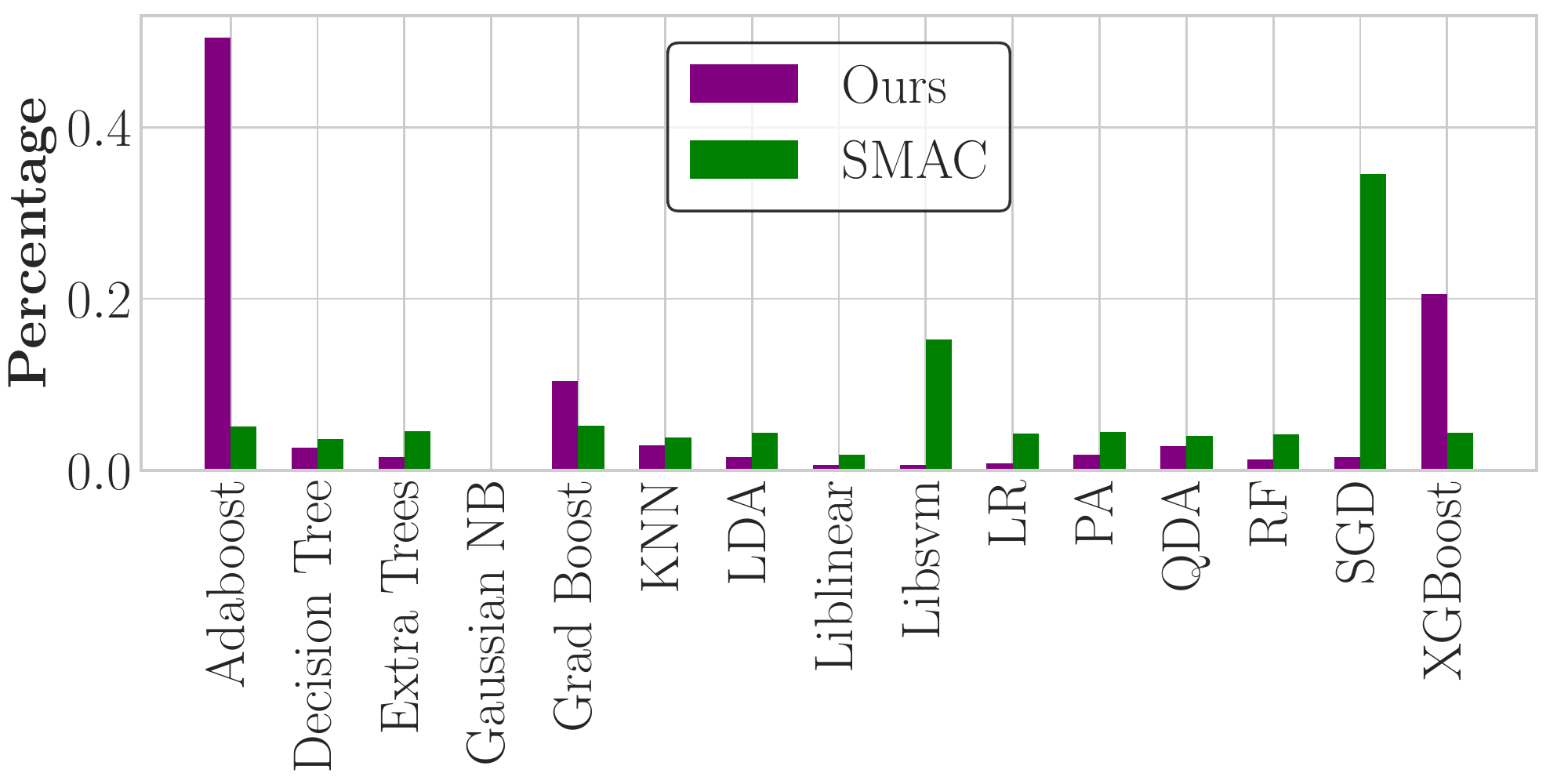}
       \includegraphics[width=0.32\columnwidth]{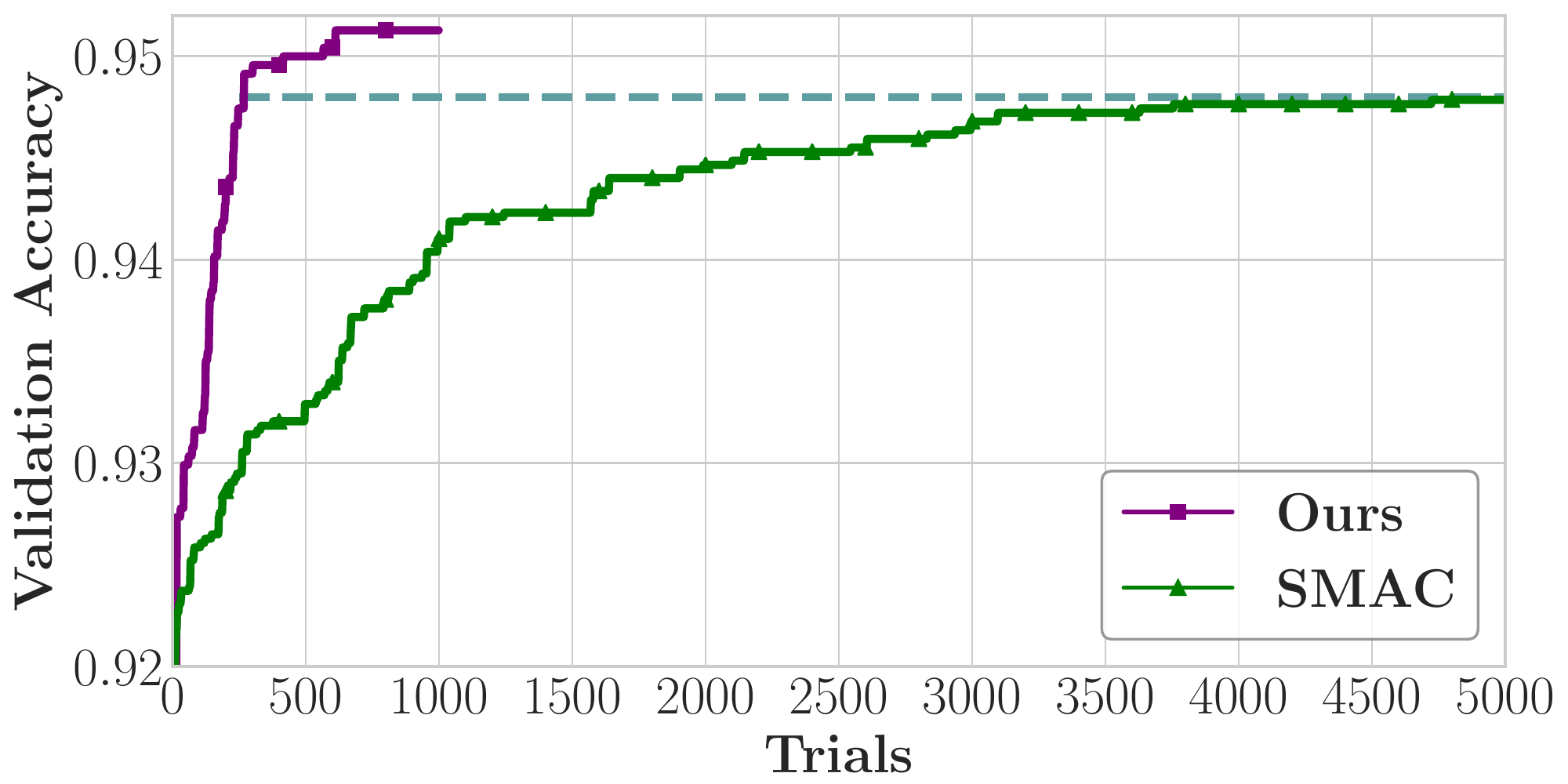}
    \end{subfigure}
    \caption{Performance comparison between BO based solutions and the proposed method on PC4 dataset.}
    \label{perf_cmp}
\end{figure*}

We compared our method with the following baselines, including the BO based methods and the traditional bandit based methods in the MAB community:
\begin{description}
    \item [AVG] The average policy that allocates the same HPO resources to each ML algorithm.
    \item [SMAC] BO based method that uses a modified random forest as the surrogate model to conduct BO.
    \item [TPE] BO based method that utilizes the tree-structured Parzen density estimators as the surrogate model.
    \item [CMAB] Bandit based method that models the stationary reward and maximizes the accumulated rewards with Thompson sampling~\cite{russo2018tutorial,DBLP:journals/corr/abs-1905-00424}.
    \item [UCB] UCB policy is used to solve the traditional MAB. 
    \item [Softmax] Softmax policy~\cite{sutton2018reinforcement} is leveraged to solve the traditional MAB.
    \item [BOHB] This method takes an adaptive strategy to conduct HPO~\cite{falkner2018bohb}.
\end{description}
In addition, to investigate its practicability and the empirical performance in the AutoML systems, we also take the following AutoML systems into account:
\begin{description}
    \item [Auto-Sklearn (ASK)] The state-of-the-art AutoML system. It utilizes the BO based solution -- SMAC to solve the CASH problem.
    \item [Hyperopt-Sklearn (HPSK)] Similar to Auto-Sklearn, it also adopts the BO based solution, and it uses TPE to conduct HPO instead.
    \item [TPOT] It leverages the genetic algorithm to solve CASH.
\end{description}

\subsubsection{CASH space, Datasets and Basic Settings}
In all experiments, the optimization space of the CASH problem is the same as the one in Auto-Sklearn.
It comprises 16 ML classification algorithms with 78 hyperparameters. 
More details about the space can be found in Appendix B of the supplemental materials.
We considered 30 classification datasets from the OpenML repositories.
These datasets are widely used in the related works~\cite{feurer2015efficient,efimova2017fast,olson2019tpot,DBLP:journals/corr/abs-1905-00424}, and the details are listed in Appendix C. 
For each run, the original dataset will be partition into three subsets: training set, validation set and test set, in the proportion of $64\%, 16\%, 20\%$.
Accuracy is used as the metric of the objective.
We repeated each method $10$ times on each dataset and reported the average accuracy.
For the sake of fairness, we assured that all compared methods use the data with the same preprocessing operations. 
That is, we processed the raw datasets with the necessary operations only (e.g., label encoder, one-hot
encoding); and we disabled the original preprocessing module in Auto-Sklearn and TPOT.
Like Auto-Sklearn and Auto-Weka, the proposed method leverages SMAC to solve the HPO problem for each ML algorithm individually.
In the following experiments, we used the initial version of our method (in Algorithm \ref{algo:algo_bandit}) by default (except when specified the concrete version).
The parameter $C$ for computing the smooth growth rate is set to $7$. 
Our method is not sensitive to the choice of $C$, and the detailed sensitivity analysis can be found in Appendix D. 

\subsubsection{More Results about the Concave Rewards}
We ran experiments on 5 datasets, and analyzed the reward functions for different ML algorithms. 
Ten figures in the supplementary materials illustrate the rewards functions for each algorithm in details.
We found that the concave behavior about the reward function is largely consistent with the result we showed in Figure~\ref{non_stationary_rewards}.

\subsection{Comparison with BO based Methods}
The empirical evaluation of BO methods shows that SMAC performs best on the benchmarks with the high-dimensional hyperparameter space, closely followed by TPE. 
In this experiment, we evaluated the performance of both the proposed method and SMAC on the CASH problem. 

\subsubsection{High-dimensional Hyperparameter Space.}
Here we demonstrated that the proposed method still works well when the hyperparameter space becomes large. 
We evaluated the following three methods on OpenML \emph{PC4} dataset with $500$ trials: average policy (AVG), SMAC and our approach (OURS).
The hyperparameter space of CASH problem is gradually augmented by adding more and more ML algorithms into the algorithm candidate $\mathcal{A}$ with $|\mathcal{A}|=K$.
The performance of each method is tested with different $K$s: $K = [1,2,4,8,12,16]$.
When $K$ equals to $1$, the hyperparameter space only includes the hyperparameters of the optimal algorithm; if $K$ is set to $16$, the hyperparameter space contains the hyperparameters of all ML algorithms and the algorithm selection hyperparameter. 
As illustrated in Table \ref{hp_dim}, SMAC suffers from the low-efficiency issue. 
With the increase of $K$, it is infeasible for BO methods to learn a surrogate model that models this huge optimization space accurately within $500$ trials. 
Consequently, the validation accuracy drops from $95.02\%$ to $93.63\%$.
In contrast, the proposed method utilizes the elimination criterion to dismiss the poor-performing algorithms from the candidate set, thus decreasing the dimension of hyperparameter space automatically. 
Hence our method still can achieve the best accuracy - $95.02\%$ when $K$ equals to $16$.

\begin{table}[htb]
\centering
\resizebox{.5\columnwidth}{!}{
  \begin{tabular}{lccc}
    \toprule
    K & AVG & SMAC & OURS  \\ 
    \midrule
    1 & 95.02 & 95.02 & 95.02 \\
    2 & 94.68 & 94.79 & 95.01 \\
    4 & 94.31 & 94.06 & 95.02 \\
    8 & 93.91 & 93.60 & 95.02 \\
    12 & 93.50 & 93.48 & 95.01 \\
    16 & 93.39 & 93.63 & 95.02 \\
    \bottomrule
  \end{tabular}
}
\caption {The validation accuracy (\%) with different $K$s in the CASH problem.}
\label{hp_dim}
\end{table}

\begin{table*}[t]
  \centering
  \resizebox{1.7\columnwidth}{!}{
  \begin{tabular}{lSSSSSSSSSSSSS}
    \toprule
    \multirow{2}{*}{\textbf{Dataset ID}} &
      \multicolumn{6}{c}{\textbf{Validation Performance (\%)}} & $ $ &
      \multicolumn{6}{c}{\textbf{Test Performance (\%)}} \\
      \cmidrule[1pt](l{2pt}r{2pt}){2-7}
      \cmidrule[1pt](l{2pt}r{2pt}){9-14} 
      & {TPE} & {S}MAC & {UCB} & {CMAB} & {SFMX} & {OURS} & 
      & {TPE} & {SMAC} & {UCB} & {CMAB} & {SFMX} & {OURS} \\
      \midrule
      1049 & 94.02 & 93.85 & 94.27 & 94.20 & 94.10 & $\bm{95.26}$ & 
        & 90.42 & 90.64 & 90.75 & 90.92 & 90.98 & $\bm{91.13}$ \\
    917 & 94.62 & 95.00 & 94.38 & 94.81 & 94.19 & $\bm{95.00}$ & 
           & 84.35 & 84.25 & 84.50 & 84.50 & 84.35 & $\bm{85.40}$ \\
    847 & 87.42 & 87.41 & 87.43 & 87.39 & 87.38 & $\bm{87.49}$ & 
    & 86.27 & 86.23 & 86.20 & 86.20 & $\bm{86.36}$ & 86.30 \\
    54 & 86.10 & 85.96 & 86.03 & 86.03 & 85.81 & $\bm{86.18}$ & 
            & 86.00 & 85.7 & 86.00 & 86.06 & 86.06 & $\bm{86.47}$ \\
    31 & 79.88 & 79.94 & 79.81 & 79.94 & 80.06 & $\bm{80.06}$ & 
             & 72.95 & 73.65 & 73.45 & 73.45 & 74.35 & $\bm{74.35}$ \\
    181 & 57.18 & 56.93 & 57.02 & 56.72 & 56.85 & $\bm{57.23}$ & 
          & $\bm{60.03}$ & 59.93 & 59.83 & 59.33 & 59.56 & 59.63 \\
    40670 & 97.76 & 97.73 & 97.90 & 97.98 & 97.84 & $\bm{98.10}$ & 
        & 96.55 & 96.60 & 96.63 & 96.69 & 96.68 & $\bm{96.77}$ \\
    40984 & 99.20 & 99.16 & 99.19 & 99.22 & 99.08 & $\bm{99.24}$ &
            & 96.80 & $\bm{97.25}$ & 96.80 & 97.14 & 97.23 & 97.14 \\
    46 & $\bm{97.63}$ & 97.48 & 97.24 & 97.32 & 97.36 & 97.44 & 
           & 95.44 & 95.27 & 95.11 & $\bm{95.56}$ & 95.44 & 95.44 \\
    772 & 60.95 & 60.20 & 60.49 & 61.03 & 60.46 & $\bm{61.20}$ & 
          & 53.19 & 53.76 & 54.06 & $\bm{54.36}$ & 54.01 & 53.85 \\
    310 & 99.00 & 98.97 & 98.97 & 98.98 & 99.00 & $\bm{99.02}$ & 
          & 98.67 & 98.71 & $\bm{98.75}$ & 98.65 & 98.67 & 98.71 \\
    40691 & 70.76 & 70.66 & 71.05 & 71.02 & 70.74 & $\bm{71.95}$ & 
                 & 66.50 & 66.09 & 65.03 & 65.97 & 66.00 & $\bm{66.66}$ \\
    1501 & 95.25 & 95.22 & 94.98 & 94.86 & 95.02 & $\bm{95.33}$ & 
            & 96.71 & 95.49 & 96.43 & 96.30 & 96.43 & $\bm{96.80}$ \\
    1557 & 67.49 & 67.52 & 67.37 & 67.58 & 67.22 & $\bm{67.85}$ & 
            & 61.71 & 62.05 & 62.09 & 61.99 & 61.99 & $\bm{62.68}$ \\
    182 & 91.99 & $\bm{92.14}$ & 92.03 & 91.95 & 91.90 & 92.04 & 
             & 91.33 & 91.40 & 91.25 & $\bm{91.52}$ & 91.32 & 91.50 \\
    823 & 98.53 & 98.50 & 98.56 & 98.54 & 98.55 & $\bm{98.60}$ & 
           & 98.08 & 98.04 & 98.01 & 98.03 & 98.04 & $\bm{98.10}$ \\
    1116 & 99.75 & 99.73 & 99.72 & 99.51 & 99.72 & $\bm{99.87}$ & 
         & 99.36 & 98.98 & 99.36 & 99.44 & 99.44 & $\bm{99.50}$ \\
    151 & 93.51 & 93.41 & 93.28 & 93.42 & 93.31 & $\bm{94.01}$ & 
         & 93.31 & 93.32 & 93.26 & 93.27 & 93.08 & $\bm{93.95}$ \\
    1430 & 85.72 & 85.69 & 85.75 & 85.62 & 85.69 & $\bm{85.85}$ & 
        & 85.09 & 85.02 & 85.13 & 85.01 & 85.06 & $\bm{85.17}$ \\
    32 & 99.55 & 99.53 & 99.45 & 99.42 & 99.47 & $\bm{99.63}$ &
              & 99.30 & 99.25 & 99.55 & 99.41 & 99.34 & $\bm{99.60}$ \\
    354 & 84.80 & 84.95 & 79.18 & 80.80 & 79.06 & $\bm{87.93}$ & 
          & 85.00 & 80.87 & 80.98 & 79.12 & 79.37 & $\bm{87.99}$ \\
    60 & 86.81 & 86.88 & 86.74 & 86.65 & 86.76 & $\bm{86.90}$ & 
             & 86.54 & 86.52 & 86.55 & 86.44 & 86.28 & $\bm{86.65}$ \\
    846 & 90.14 & 90.12 & 90.15 & 90.05 & 90.16 & $\bm{90.19}$ & 
 & 89.01 & 89.00 & 88.74 & 88.90 & 89.04 & $\bm{89.07}$ \\
    28 & 98.85 & 98.81 & 98.77 & 98.59 & 98.78 & $\bm{98.87}$ &
              & 98.84 & 98.73 & 98.84 & 98.84 & 98.81 & $\bm{98.85}$ \\
    1471 & 97.99 & 97.93 & 97.84 & 97.50 & 97.93 & $\bm{98.28}$ & 
            & 97.75 & 97.38 & $\bm{98.08}$ & 97.83 & 97.74 & 97.61 \\
    9976 & $\bm{87.02}$ & $\bm{87.02}$ & 86.54 & 86.97 & 85.82 & 86.83 & 
            & 85.85 & $\bm{86.62}$ & 85.65 & 85.46 & 85.58 & 86.60 \\
    23512 & 72.96 & 73.12 & 72.96 & 72.80 & 72.90 & $\bm{73.20}$ & 
 & 72.60 & 72.29 & 72.55 & 72.46 & 72.60 & $\bm{72.86}$ \\
    41082 & 97.89 & 97.74 & 97.65 & 97.10 & 97.74 & $\bm{98.10}$ & 
         & 97.54 & 97.10 & 97.56 & 97.54 & 97.55 & $\bm{97.62}$ \\
    389 & $\bm{87.73}$ & 86.60 & 86.80 & 86.66 & 86.60 & 87.70 & 
            & $\bm{87.56}$ & 86.37 & 86.98 & 87.22 & 87.38 & 87.51 \\
    184 & 89.33 & 89.12 & 89.23 & 89.17 & 89.19 & $\bm{89.65}$ & 
          & 88.34 & 88.20 & 88.21 & 88.18 & 88.22 & $\bm{88.78}$\\
    \bottomrule
  \end{tabular}
  }
  \caption{Average validation accuracy and test accuracy for all considered methods on $30$ OpenML datasets.}
  \label{exp_results}
\end{table*}

\subsubsection{Resource Allocation}
Figure \ref{perf_cmp} (a) depicts the validation accuracy of three methods across trials, where $500$ trials are used to solve the CASH problem with $K=16$.
In the first $100$ trials, SMAC and the proposed method behave similarly, and both of them explore the performance distribution over the optimization space. 
Then our method starts to identify and dismiss the poor-performing algorithms by leveraging the known HPO results. 
More resources (trials) are allocated to the more promising algorithms, and this procedure brings significant performance improvement. 
Due to the huge hyperparameter space, SMAC cannot model the performance for each ML algorithm effectively. 
Therefore, its performance improves very slowly,
and it is even worse than the average policy.
To further compare our method with SMAC, Figure \ref{perf_cmp} (b) illustrates their percentages of the HPO trials for each ML algorithm respectively. 
In this problem (dataset), Adaboost is the optimal algorithm. 
As can be seen, our method identifies and terminates $13$ unpromising ML algorithms by using $20\%$ trials. 
Another $30\%$ of trials are used to dismiss the left two algorithms that have a near-optimal performance. 
Almost $50\%$ of trials are spent on tuning the optimal algorithm -- Adaboost. 
In contrast, most of the trials in SMAC are used to tune the poor-performing algorithms. 

\subsubsection{Speedups}
We evaluated the achievable speedups of our method against the baseline - SMAC on 10 OpenML datasets.
Continued with the previous settings, $5000$ trials in total are given to SMAC.
The speedup is measured with the number of trials (\#) that each method needs to reach the same validation accuracy (\%). 
Table \ref{speedups_vs_bo} depicts the speedup results. 
As can be seen, our method is more efficient than SMAC in terms of the number of trials one needs to reach the same validation accuracy.
To derive a more clear illustration about this, we plotted the validation accuracy curve of these two methods across trials on the PC4 dataset.
As shown in Figure \ref{perf_cmp} (c), the final validation accuracy of SMAC is still worse than the one that our approach achieves within $250$ trials. 
The empirical results demonstrate that the proposed method can outperform the existing CASH algorithm - SMAC by over an order of magnitude speedups. 
\begin{table}[t]
\centering
\resizebox{.85\columnwidth}{!}{
  \begin{tabular}{lcccc}
    \toprule
    Dataset ID & Val Acc & \#SMAC & \#OURS & Speedups  \\ 
    \midrule
    1049  & 94.81 & 5000 & 250 & 20.0x \\
    40691 & 71.38 & 5000 & 395 & 12.7x \\
    40670 & 97.86 & 5000 & 230 & 21.7x \\
    847   & 87.48 & 5000 & 480 & 10.4x \\
    32    & 99.61 & 5000 & 450 & 11.1x \\
    151   & 93.94 & 5000 & 350 & 14.3x \\
    184   & 89.63 & 4000 & 500 & 8.00x \\
    354   & 87.53 & 5000 & 427 & 11.7x \\
    1471  & 98.20 & 5000 & 500 & 10.0x \\
    41082 & 98.10 & 3000 & 500 & 6.20x \\
    \midrule
    \textbf{Average}   & - & -    & -   & \textbf{12.6x} \\
    \bottomrule
  \end{tabular}
}
\caption {Speedup results on 10 OpenML datasets.}
\label{speedups_vs_bo}
\end{table}

\subsection{Comparison with All Considered Methods}
In this experiment, we compared the proposed method with all considered baselines in terms of two perspectives: 1) final accuracy, and 2) the efficiency, i.e., the number of trials one needs to reach the same validation accuracy.
In the first part, each method is given $500$ trials, and the average accuracy across $10$ runs is reported.
Table \ref{exp_results} lists both the average validation accuracy and the average test accuracy on 30 OpenML datasets.
In order to evaluate the generalization of the corresponding model, we also compared the accuracy on the test set.
As can be seen, the proposed method achieves the best validation accuracy on $26$ out of $30$ datasets, and it also reaches the highest test accuracy on $20$ out of $30$ datasets. 
This gives that the ML models obtained by our method generalize well.
Although our method does not get the highest accuracy on a few datasets, its result is very close to the best one.
It is worth noting that, on most datasets, our method outperforms both the existing bandit-based methods (CMAB, UCB, and Softmax) and BO-based methods in terms of the final accuracy in solving the CASH problem. 

In the second part, we took another two related works into consideration: Heidari et al.~\cite{heidari2016tight} and BOHB~\cite{falkner2018bohb}. 
First we ran these two methods on 10 datasets with 500 trials, and the result is reported in Table \ref{speedups_vs_all}. 
Although Heidari et al. (2016) leverage the concave reward function, this method cannot outperform the solution found by our approach because it tries to maximize the accumulated rewards. 
As mentioned previously, the objective in CASH focuses more on identifying the optimal arm, instead of optimizing the accumulated rewards.
Similar to our approach, BOHB adopts an adaptive mechanism to conduct hyperparameter optimization. 
The reason why this method cannot beat our method is that it does not use the structure information about the concave rewards in CASH. 
By contrast, our method, with the Rising Bandits, absorbs the advantages of these two kinds of methods, and avoids their drawbacks successfully.
Furthermore, similar to the last section about speedups, we gave the baseline - BOHB enough trials, enabling it to reach the same validation accuracy that our method gets within 500 trials (that is, the fourth column in Table \ref{speedups_vs_all}). 
Finally, we obtained the speedups against BOHB, and illustrated the result in Table \ref{speedups_vs_all}. 
It exhibits that the CASH-oriented Rising Bandits are more efficient than the existing adaptive method in solving the CASH problem.

\begin{table}[t]
\centering
\resizebox{.9\columnwidth}{!}{
  \begin{tabular}{lcccc}
    \toprule
    Dataset ID & Heidari et al & BOHB & OURS & Speedups against BOHB\\ 
    \midrule
    1049  & 94.26 & 94.31 & $\bm{95.25}$ & 8.0x \\
    40691 & 71.05 & 71.06 & $\bm{71.95}$ & 7.5x \\
    40670 & 97.82 & 97.79 & $\bm{98.10}$ & 8.6x \\
    847   & 87.35 & 87.42 & $\bm{87.49}$ & 2.4x \\
    32    & 99.42 & 99.52 & $\bm{99.63}$ & 3.9x \\
    151   & 93.25 & 93.64 & $\bm{94.01}$ & 5.3x \\
    184   & 89.18 & 89.40 & $\bm{89.65}$ & 4.5x \\
    354   & 80.79 & 85.18 & $\bm{87.90}$ & 15.7x \\
    1471  & 97.99 & 97.87 & $\bm{98.28}$ & 5.7x \\
    41082 & 97.69 & 97.96 & $\bm{98.12}$ & 3.5x \\
    \bottomrule
  \end{tabular}
}
\caption {Average validation accuracy (\%) and speedups compared with the considered methods.}
\label{speedups_vs_all}
\end{table}

\subsection{Comparison with AutoML Systems}
To investigate the practicality and effectiveness of our method in the AutoML systems, we implemented the proposed method based on the components of Auto-Sklearn and compared it with three popular AutoML systems. 
Each system is given $2$ hours, and the average test accuracy across $10$ runs is reported.
The cost-aware variant of our method is used to solve the CASH problems.
Because the three AutoML systems do not take the evaluation cost into account,
they only optimize the performance, instead of optimizing both efficiency and performance together.
As a result, given a limited time, these AutoML systems suffer from the low-efficiency problem.
The empirical results in Table \ref{ens_result} demonstrate that the proposed method is more efficient than the existing AutoML systems on the $12$ OpenML datasets. 

\begin{table}[t]
\centering
\resizebox{.8\columnwidth}{!}{
  \begin{tabular}{lccccc}
    \toprule
    Dataset & ASK & HPSK & TPOT & OURS \\ 
    \midrule
    AMAZON & 72.33 & 73.67 & 75.45 & $\bm{82.60}$ \\
    POKER & 84.91 & 84.83 & 81.59 & $\bm{85.92}$ \\
    WINE & 65.69 & 65.61 & 65.54 & $\bm{66.76}$ \\
    FBIS-WC & 86.17 & 86.21 & 86.61 & $\bm{87.30}$ \\
    OPTDIGITS & 98.79 & 98.78 & $\bm{99.16}$ & 99.10 \\
    SEMEION & 96.55 & 96.57 & 96.36 & $\bm{96.99}$ \\
    HIGGS & 71.98 & 71.81 & 71.58 & $\bm{72.20}$ \\
    PC4 & 91.16 & 91.07 & 90.94 & $\bm{91.21}$ \\
    USPS & 96.42 & 96.57 & 97.47 & $\bm{97.66}$ \\
    MUSK & 99.29 & 99.20 & 99.63 & $\bm{99.73}$ \\
    ELEVATORS & 88.64 & 88.71 & 88.86 & $\bm{89.01}$ \\
    ELECTRICITY & 93.11 & 92.98 & 90.16 & $\bm{93.84}$ \\
    \bottomrule
  \end{tabular}
  }
\caption {Average test accuracy (\%) of compared AutoML systems on $12$ OpenML datasets.}
\label{ens_result}
\end{table}

\section{Conclusion}
In this paper, we proposed an alternating optimization framework to accelerate the CASH problem, where a novel MAB variant is introduced to conduct algorithm selection and the Bayesian optimization methods are used to conduct HPO for each ML algorithm individually.
Moreover, we presented an online algorithm to solve the Rising Bandits problem with provably theoretical guarantees.
We evaluated the performance of the proposed method on a number of AutoML tasks and demonstrated its superiority over the competitive baselines. 
In the future work, we plan to leverage the meta-learning techniques to speed up the CASH problem.

\section{Acknowledgments}
This work is supported by the National Key Research and Development Program of China (No.2018YFB1004403), NSFC (No.61832001, 61702015, 61702016, 61572039), Beijing Academy of Artificial Intelligence (BAAI), and Alibaba-PKU joint program. Jiawei Jiang is the corresponding author.

\bibliographystyle{aaai}
\bibliography{reference}
\end{document}